\documentclass{article}
\usepackage{ijcai19}

\usepackage{times}
\usepackage{soul}
\usepackage{url}
\usepackage[hidelinks]{hyperref}
\usepackage[utf8]{inputenc}
\usepackage[small]{caption}
\usepackage{graphicx}
\usepackage{amsmath}
\usepackage{booktabs}
\usepackage{dblfloatfix}

\usepackage{amssymb}
\usepackage{xcolor, multirow}
\usepackage{tikz} 
\usetikzlibrary{fit,positioning}
\usepackage{caption}
\usepackage{subcaption}
\usepackage{tabularx}

\newcommand{\X}{\mathbf{X}}

\newcommand{\x}{\mathbf{x}}
\newcommand{\Y}{\mathbf{Y}}
\newcommand{\y}{\mathbf{y}}
\newcommand{\f}{\mathbf{f}}
\newcommand{\F}{\mathbf{F}}

\newcommand{\W}{\mathbf{W}}

\newcommand{\R}{\mathbb{R}}

\newcommand{\I}{\mathbf{I}}

\newcommand{\K}{\mathbf{K}}

\newcommand{\N}{\mathcal{N}}

\newcommand{\ud}{\mathrm{d}}

\title{Scalable Bayesian Non-linear Matrix Completion}

\author{
Xiangju Qin\footnote{Contact Author. Currently at Institute for Molecular Medicine Finland (FIMM), University of Helsinki.}\and
Paul Blomstedt\And
Samuel Kaski
\affiliations
Department of Computer Science, Helsinki Institute for Information Technology HIIT, \\ 
Aalto University, 00076 Espoo, Finland\\
\emails
xiangju.qin@helsinki.fi,
paul.blomstedt@aalto.fi,
samuel.kaski@aalto.fi
}

\begin{document}

\maketitle

\begin{abstract}
Matrix completion aims to predict missing elements in a partially observed data matrix which in typical applications, such as collaborative filtering, is large and extremely sparsely observed. A standard solution is matrix factorization, which predicts unobserved entries as linear combinations of latent variables. We generalize to non-linear combinations in massive-scale matrices. Bayesian approaches have been proven beneficial in linear matrix completion, but not applied in the more general non-linear case, due to limited scalability. We introduce a Bayesian non-linear matrix completion algorithm, which is based on a recent Bayesian formulation of Gaussian process latent variable models. To solve the challenges regarding scalability and computation, we propose a data-parallel distributed computational  approach with a restricted communication scheme.  We evaluate our method on challenging out-of-matrix prediction tasks using both simulated and real-world data.
\end{abstract}

\section{Introduction}

In matrix completion---one of the most widely used approaches for collaborative filtering---the objective is to predict missing elements of a partially observed data matrix. Such problems are often characterized by large and extremely sparsely observed data sets. The classic linear solution to the problem is to find a factorization of the data matrix $\Y\in \R^{N\times D}$ as a product of latent variables $\X\in \R^{N\times K}$ and weights $\W\in \R^{D\times K}$ ($K\ll N,D$), from which elements of $\Y$ can be predicted as $\Y \approx \X \W^T$.
Probabilistic matrix factorization (PMF) \cite{Salakhutdinov+Mnih:2008a}, formulates the problem as a probabilistic model, regularized by placing priors on $\X$ and $\W$, and finds the solution as a maximum a posteriori (MAP) estimate of these matrices. Fully Bayesian matrix factorization \cite{Salakhutdinov+Mnih:2008b} expands this model by further placing priors on model hyperparameters, and marginalizing these along with $\X$ and $\W$. Bayesian matrix factorization brings the advantages of automatic complexity control and better robustness to overfitting. Moreover, the solution comes with an uncertainty estimate, which is useful when the completed matrix is used for decision making. For instance, sparsely observed drug-target interaction matrices are used for deciding which interactions to measure next.

Lawrence and Urtasun \shortcite{Lawrence+Urtasun:2009} generalized PMF using a Gaussian process latent variable model (GP-LVM) formulation, where the relationship between $\X$ and $\Y$ is given by $\Y \approx \f(\X)$, with a GP-prior placed over $\f$. The $\X$ is optimized to find its MAP solution.
Note that this formulation also subsumes the linear model as a special case. 
Subsequently, a variational inference framework for fully Bayesian GP-LVM has been developed \cite{Damianou+others:2016,Titsias+Lawrence:2010}, building on sparse GP approximations \cite{Quinonero-Candela+Rasmussed:2005,Snelson+Ghahramani:2006}. It parametrizes the covariance matrix implied by the GP kernel using a set of $M\ll N$ auxiliary inducing variables. While Bayesian GP-LVM has been successfully used for dimensionality reduction and extracting latent representations, less consideration has been given to its applicability in matrix completion tasks with extremely sparse data. Computationally, this is a much more demanding problem, because the variational updates have to be performed separately for each dimension of the data matrix, instead of being performed as a single operation. 

Existing approaches for scaling up Bayesian GP-LVM make use of the insight that, conditional on the inducing variables, the data points can be decoupled for parallel computations. In this line of work,  Gal et al. \shortcite{Gal+others:2014}  introduced a distributed version of Bayesian GP-LVM.  Dai et al. \shortcite{Dai+others:2014} proposed a similar framework, additionally using GPU acceleration to speed up local computations. Neither of the works demonstrated learning of latent variable models beyond moderately-sized data, nor have they been implemented for sparse matrices, which is needed for the problems considered in this paper. 
More importantly, current distributed solutions require the worker nodes to communicate with the master node in every iteration, which leads to an accumulating communication overhead as the number of worker units increased with the size of the problem. Vander Aa et al. \shortcite{VanderAa+others:2016} reported such a phenomenon for their distributed MPI implementation of Bayesian linear matrix factorization.
Finally, our experience indicates that existing distributed implementations may suffer from high memory consumption. 

For GP-regression models, with $\X$ observed, Deisenroth and Ng \shortcite{Deisenroth+Ng:2015} proposed a framework with particularly good scaling properties and efficient use of memory. This framework utilizes a product-of-GP-experts (PoE) formulation \cite{Cao+others:2014,Ng+Deisenroth:2014,Tresp:2000}, which makes predictions using a product of independent local models, each operating on a subset of the data. These types of approximations are amenable to embarrassingly parallel computations, 
and can therefore be scaled up to arbitrarily large data sets, at least in principle. 
However, a direct application of PoE for nonlinear matrix completion may not produce satisfactory predictions for two reasons.
First, since the target matrix is very sparsely observed, 
each local model has very limited information to learn an informative model without sharing information. 
Second, while local models could be combined into larger models to improve predictive performance, this is hindered by the general non-uniqueness of the latent variables in latent variable models.
%a direct application of PoE for nonlinear matrix completion may not produce satisfactory predictions without a mechanism for sharing information between local models. 

%we need a mechanism to encourage the local predictions to share information.
%the general non-uniqueness of the latent variables in latent-variable models may affect the predictive performance when aggregating 
%However, a direct application of PoE for GP-LVMs is hindered by the general non-uniqueness of the latent variables in latent variable models. This property makes the aggregation of predictions by local models difficult to carry out in a meaningful way. 

In this work, we propose a distributed computational strategy which is almost as communication-efficient as embarrassingly parallel computation, but 
%avoids the problem of non-uniqueness. 
enables local models to share information and avoids the problem of non-uniqueness in aggregating local models.  
In a nutshell, one data subset is processed first, and the rest of the embarrassingly parallel computations are conditioned on the result. A similar idea was recently presented by \cite{Qin+others:2019} for Bayesian linear matrix completion. 
%For a comparison of available implementations of non-linear matrix completion and large-scale Bayesian GP-LVM, see Table~\ref{tab:comparison}.
% 
The remainder of the paper proceeds as follows: In Section~\ref{sec:gp-lvm} we first provide a brief review of GP-LVMs. Then, in Section~\ref{sec:framewok}, we present our framework for scalable Bayesian non-linear matrix completion. An empirical evaluation of the method, using simulations and a benchmark dataset, is given in Section~\ref{sec:experiments}. The paper ends with conclusions in Section~\ref{sec:conclusion}.

%%%%%%%% commented block %%%%%%%%%%%%%%%%%%%%%%%%%%%%
\iffalse

\begin{table*}[!htbp]
\caption{Comparison of implementations for non-linear matrix completion and large-scale Bayesian GP-LVM.}\label{tab:comparison}
\begin{tabularx}{\textwidth}{|l|X|X|X|X|}
\hline 
 & Sparse data & Multi-view & Distributed computation & Communication-efficient\\ 
\hline 
Non-lin. PMF \cite{Lawrence+Urtasun:2009} & \checkmark &  &  & N/A\\ 
\hline 
GPy \cite{Dai+others:2014} &  & \checkmark & \checkmark &\\ 
\hline 
GPflow \cite{Matthews+others:2017} &  &  & \checkmark &\\ 
\hline 
This paper & \checkmark & \checkmark & \checkmark & \checkmark\\ 
\hline 
\end{tabularx}
\end{table*}

\fi
%%%%%%%%%%%%%%%%%%%%%%%%%%%%%%%%%%%%%%%%%%%%%%%%%%%%%%

\section{Gaussian Process Latent Variable Models}\label{sec:gp-lvm}

A Gaussian process latent variable model (GP-LVM) \cite{Lawrence:2005} can be constructed from a non-linear multi-output regression model,
\begin{align*}
p(\Y|\F,\X,\sigma^2) &= \prod_{d=1}^D p(\y_{:,d}|\f_{:,d},\X,\sigma^2)\\  
&= \prod_{d=1}^D \prod_{n=1}^N \mathcal{N}\left(y_{n,d}|f_d(\mathbf{x}_{n,:}),\sigma^2\right),
\end{align*}
by placing a GP prior over the unknown functions $f_1\ldots,f_d$. Integrating over the space of functions with respect to a zero-mean GP then yields the likelihood as
\begin{align}
\label{eq:gplvm_likelihood}
p(\Y | \X, \boldsymbol{\theta}) %&= \int p(\Y|\F) p(\F|\X,\boldsymbol{\theta}) \ud\F \\
&= \prod_{d=1}^D \int \mathcal{N}(\y_{:,d}|\f_{:,d},\sigma^2 \I)\mathcal{N}(\f_{:,d}|\boldsymbol{0}, \K)\ud \f_{:,d} \nonumber 
 \\ 
	&=\prod_{d=1}^{D} \N( \y_{:,d} | \boldsymbol{0}, \K + \sigma^2 \I ),
\end{align}
where $\K$ is an $N \times N$ kernel matrix defined by a GP covariance function $k(\mathbf{x}_{s,:},\mathbf{x}_{t,:})$. We use $\boldsymbol{\theta}$ to collectively denote all parameters, including the noise variance $\sigma^2$ and the parameters of the covariance function.

When values are missing, as is the case in matrix completion, each factor of the likelihood (\ref{eq:gplvm_likelihood}) will only account for observed elements, thus we have  
\begin{equation}\label{eq:gplvm_likelihood_missing_data}
p(\Y | \X, \boldsymbol{\theta}) = \prod_{d=1}^{D} \N( \y_{\textbf{n}_d,d} | \boldsymbol{0}, \K + \sigma^2 \I ),
\end{equation}
where $\textbf{n}_d$ denotes the set of indices of observed elements in column $d$. Furthermore, the symmetric matrices $\K$ and $\I$ will only include rows and columns corresponding to the indices $\textbf{n}_d$.

\subsection{Bayesian GP-LVM}\label{sec:BayesianGPLVM}

For Bayesian GP-LVM, we complement the likelihood (\ref{eq:gplvm_likelihood_missing_data}) by placing a prior on $\X$. A standard choice is to set 
\begin{equation}\label{eq:X_prior}
p(\X) = \prod_{n=1}^N \mathcal{N}(\x_{n,:}|\boldsymbol{0},\I). 
\end{equation}
The marginal likelihood of Bayesian GP-LVM is obtained by integrating the model with respect to $p(\X)$:
\begin{equation} \label{eq:marginal_likelihood}
p(\Y| \boldsymbol{\theta}) = \int p(\Y|\X, \boldsymbol{\theta})p(\X)\ud\X.
\end{equation} 
While this operation is intractable in general, Titsias and Lawrence \shortcite{Titsias+Lawrence:2010} introduced a variational framework, which leads to a tractable lower bound, 
\begin{align}\label{eq:lower_bound}
F(q) &= \int q(\X) \log\frac{p(\Y|\X, \boldsymbol{\theta})p(\X)}{q(\X)}\ud\X  \\
     &= \int q(\X) \log p(\Y|\X, \boldsymbol{\theta}) \ud\X - \int q(\X) \log\frac{q(\X)}{p(\X)}\ud\X \nonumber \\
     &= \int q(\X) \log p(\Y|\X, \boldsymbol{\theta}) \ud\X - \mathrm{KL}\left(q(\X)||p(\X)\right)\nonumber,
\end{align}
on the log of the marginal likelihood (\ref{eq:marginal_likelihood}). 
For a detailed treatment of the framework, see \cite{Damianou+others:2016}.
As a by-product of optimizing the lower bound (\ref{eq:lower_bound}), we get a variational approximation $q(\X)$ to the posterior $p(\X|\Y)\propto  p(\Y|\X)p(\X)$, for which we assume a factorized Gaussian form 
\begin{align}\label{eq:variational_post}
    q(\X) = \prod_{n=1}^N \mathcal{N}(\x_{n,:}|\boldsymbol{\mu}_n,S_n). 
\end{align}
In our current work, we use this approximation to share information between parallel computations (see Section~\ref{sec:propagation}).

\subsection{Extension to Multi-view Settings}

Manifold relevance determination (MRD) \cite{Damianou+others:2012} extends GP-LVM to a multi-view setting by reformulating the likelihood (\ref{eq:gplvm_likelihood}) as 
$
\prod_{v\in\mathcal{V}} p(\Y^{v}|\X,\boldsymbol{\theta}^{v})$, 
where the elements of $\mathcal{V}$ index the \emph{data views}, i.e. matrices conditionally independent given a single latent matrix $\X$.
In matrix completion problems, one of the views is typically the target in which prediction is carried out, while the other views constitute side-data. % to support prediction in the target view. 
%When the target view has completely unobserved (or new) rows or columns, predictions are effectively done `outside' of the 
%observed matrix, with the help of observed data in the side-views. This setting is referred to as \emph{out-of-matrix} prediction.
When predicting values in completely unobserved (or new) rows or columns in the target, predictions are effectively done `outside' of the 
observed matrix. This can be done with the help of observed data in the side-views, and is referred to as \emph{out-of-matrix} prediction.

\begin{figure*}[!htb]
\begin{center}
\includegraphics[width = 0.70\textwidth]{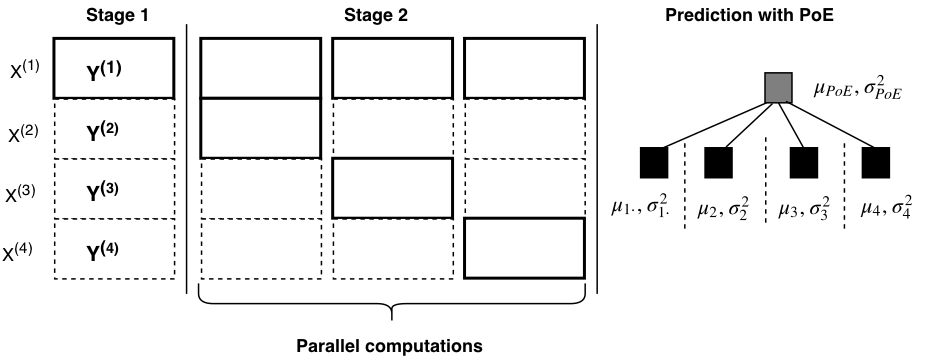}
\end{center}
\caption{Overview of scalable Bayesian non-linear matrix completion. In the learning phase (left panel), a large matrix is partitioned into 4x1 subsets, which are processed in two stages: in the initial stage only one subset is processed (box with solid borders), after which the remaining subsets are processed in parallel, each coupled with the first subset using incremental learning. The prediction phase (right panel) uses a product of experts, aggregating the means $\mu_i$ and variances $\sigma^2_i$ of local experts into a joint Gaussian prediction with $\mu_{\text{PoE}}$ and $\sigma^2_{\text{PoE}}$.}\label{fig:overview}
\end{figure*}

\section{Scalable Matrix Completion Using Bayesian GP-LVM}\label{sec:framewok}

This section presents a computational strategy, which enables Bayesian GP-LVM to be scaled up for large-scale matrix completion problems using a product of experts (PoE). 
In brief,  we first partition the observation matrix $\Y$ into $I$ disjoint subsets $\Y^{(1)},\Y^{(2)}\ldots, \Y^{(I)}$ for parallel processing. To avoid problems resulting from the unidentifiability of latent variable models, we couple the subsets as follows: in an initial stage only one subset is processed; in the second stage, the remaining subsets are processed in parallel using incremental learning (Section~\ref{sec:propagation}). For each subset, we use an implementation of the variational inference framework for Bayesian GP-LVM \cite{Titsias+Lawrence:2010}. Finally, with a set of independently learned Bayesian GP-LVMs, we use PoE to predict unobserved matrix elements (Section~\ref{sec:PoE}). 

The proposed strategy is summarized in Figure~\ref{fig:overview}.  In Section~\ref{sec:intermediate_aggregation}, we present a further variant of the method, which uses intermediate aggregation of the submodels, prior to PoE, to improve predictive performance. The scalability of our method is briefly discussed in Section~\ref{sec:scalability}.

%\subsection{Propagating information for parallel computations} \label{sec:propagation}
\subsection{Coupling Parallel Inferences Using Incremental Learning}\label{sec:propagation}

To couple the parallel inferences over subsets in the second stage of our method, we use a probabilistic variant of the incremental learning algorithm introduced by \cite{Yao+others:2011} for online learning of (non-Bayesian) non-linear latent variable models.
Let $\Y^{(1)}$ be the submatrix processed in the initial stage. Furthermore, denote by $\Y_{\text{aug}}^{(i)} = [\Y^{(1)},\Y^{(i)}]\in \R^{(N_1+N_i)\times D}$, $i=2,\ldots,I$, the combined submatrix obtained by augmenting $\Y^{(i)}$ with $\Y^{(1)}$. The corresponding combined latent matrix is denoted by $\X_{\text{aug}}^{(i)} = [\X^{(1)},\X^{(i)}]\in \R^{(N_1+N_i)\times K}$.

The objective of incremental learning is to learn the joint latent matrix $\X_{\text{aug}}^{(i)}$ without extensive relearning of $\X^{(1)}$, while still allowing it to be updated.
When learning $\X_{\text{aug}}^{(i)}$, Yao et al. \shortcite{Yao+others:2011} added a regularizer to the log-likelihood to prevent $\X^{(1)}$ from deviating too much from its initial estimate, and to speed up learning. 
The original incremental learning algorithm used the Frobenius norm $\| \X^{(1)} - \hat{\X}^{(1)}\|_{F}^2$ to regularize the updated estimate of $\X^{(1)}$.
In our current work, we use the KL-divergence $\mathrm{KL}\left(q(\X^{(1)})||\hat{q}(\X^{(1)})\right)$ to ensure that the updated variational posterior approximation $q(\X^{(1)})$ remains close to the initial approximation $\hat{q}(\X^{(1)})$.
For $\X^{(i)}$, we use the default prior given in Equation~(\ref{eq:X_prior}).
Thus, the variational inference for the incremental learning of Bayesian GP-LVM follows the procedure introduced in Section~\ref{sec:BayesianGPLVM}, with the KL terms in the lower bound of Eq.~(\ref{eq:lower_bound}) taking the following form:
\begin{align*}\label{eq:kl_variational_prior}
\mathrm{KL}\left(q(\X_{\text{aug}}^{(i)})||p(\X_{\text{aug}}^{(i)})\right) =~&\mathrm{KL}\left(q(\X^{(1)})||\hat{q}(\X^{(1)})\right) +\nonumber \\
     \phantom{=~}&\mathrm{KL}\left(q(\X^{(i)})||p(\X^{(i)})\right).
\end{align*}

For the augmented subsets, the inducing points and kernel hyperparameters are initialized to values obtained in the initial stage. 
For initialization of latent variables, we use posterior means for $\X^{(1)}$ and nearest neighbors for $\X^{(i)}$ \cite{Yao+others:2011}.

\subsection{Prediction with Product of Experts}\label{sec:PoE}

%Explain PoE and the required modification: since we do PoE on the augmented subsets, we need to account for the fact that we used first subset multiple times.
Product of experts (PoE) prediction for Gaussian process regression \cite{Deisenroth+Ng:2015} uses the simple idea of combining predictions made by independent GP-models (i.e. `experts') as a product:
%The basic PoE model for GP-regression is formulated as:
%\[
\begin{equation}\label{eq:poe_gp_regression}
p(y_*|\x_*, \X) = \prod_{i=1}^I p_i\left(y_*|\x_*, \X^{(i)}\right),
\end{equation}
%\]
where $\x_*$ is a given test input and $y_*$ is the corresponding output value to be predicted.  
Under this model, a prediction is proportional to a Gaussian with parameters
\begin{align*}
\mu_*^{\text{poe}} &= (\sigma_*^{\text{poe}})^2 \sum_{i=1}^I \sigma_i^{-2}(\x_*)\mu_i(\x_*),\\
(\sigma_*^{\text{poe}})^{-2} &= \sum_{i=1}^I \sigma_i^{-2}(\x_*).
\end{align*}

With latent variable models, the essential difference to the above is that the inputs are unobserved and must therefore be inferred. 
In matrix completion, we wish to predict the missing part of a partially observed test point $\y_*=(\y^{O}_*, \y^{U}_*) \in \R^{D}$, where $\y^{O}_*$ are the observed elements (or side views) and $\y^{U}_*$ are the missing values (or the unobserved target view) to be reconstructed. The prediction task can be finished in two steps. First, we infer the latent input $\x_*$ of the test point, which involves 
% optimizing the parameters ($\boldsymbol{\mu}_*,S_*$) of the variational distribution $q(\x_*)$ by 
maximizing the variational lower bound on the marginal likelihood 
\[
p(\y^{O}_*,\Y)=\int p(\y^{O}_*,\Y|\X, \x_*,\boldsymbol{\theta})p(\X,\x_*)\ud\X\ud\x_*
\]
to obtain the approximate posterior $q(\X,\x_*)=q(\X)q(\x_*)$.
The lower bound has the same form as the learning objective function in Equation (\ref{eq:lower_bound}), but for its maximization, the variational distribution $q(\X)$ over latent variables for training data and parameters 
$\boldsymbol{\theta}$ remains fixed during test time. 
%After inferring the latent test input $\x_*$, we follow \cite{Damianou+others:2016} and take the standard GP prediction approach which can take into account uncertain inputs 
%($q(\x_*)$ is a distribution) to make predictions for $\y^{U}_*$.
After obtaining $q(\x_*)$, making predictions for $\y^{U}_*$ is approached as GP prediction with uncertain inputs \cite{Damianou+others:2016,Girard+others:2003}.
In our distributed setting, the experts in PoE correspond to submodels learned from the augmented training subsets formed in the incremental learning phase. To correct for the initial subset $\Y^{(1)}$ being used in $I-1$ training sets, we formulate a corrected PoE as follows:
%\begin{align*}
%& p(\y_*|\Y, \boldsymbol{\theta}) = \\ 
%& \phantom{p(\y_*}\prod_{i=2}^I\left[ p_{1}(\y_*|\Y^{(1)}, \boldsymbol{\theta})^{\frac{2-I}{I-1}} p_i(\y_*|\Y^{(i)}, \boldsymbol{\theta})\right],
%\end{align*}
%\begin{align*}
%& p(\y_*|\x_*,\Y, \boldsymbol{\theta}) = p_{1}(\y_*|\x^{(1)}_*,\Y^{(1)}, \boldsymbol{\theta}) \\ 
%& \phantom{p(\y_*}\prod_{i=2}^I\left[ p_{1}(\y_*|\x^{(1)}_*,\Y^{(1)}, \boldsymbol{\theta})^{-1} p_i(\y_*|\boldsymbol{x}^{(i)}_*,\Y^{(i)}_{aug}, \boldsymbol{\theta})\right],
%\end{align*}
%\begin{align*}
%& p(\y_*|\x_*,\Y, \boldsymbol{\theta}) = \\ 
%& \phantom{p(\y_*}p_{1}(\y_*|\x^{(1)}_*,\Y^{(1)}, \boldsymbol{\theta})\prod_{i=2}^I\left[ p_{i}(\y_*|\x^{(1)}_*,\Y^{(1)}, \boldsymbol{\theta})^{-1} p_i(\y_*|\boldsymbol{x}^{(i)}_{*},\Y_{\text{aug}}^{(i)}, \boldsymbol{\theta})\right],
%\end{align*}
\begin{align*}
 p(\y_*|\Y, \boldsymbol{\theta}) &= 
p_{1}(\y_*|\Y^{(1)}, \boldsymbol{\theta}^{(1)}) \times \\
& \prod_{i=2}^I\left[ p_i(\y_*|\Y_{\text{aug}}^{(i)}, \boldsymbol{\theta}^{(i)})p_{1}(\y_*|\Y^{(1)}, \boldsymbol{\theta}^{(1)})^{-1} \right],
\end{align*}
%where %the index $1_i$ refers to the repeated part of the $i$th training set, 
%the superscript $i$ in the latent test input $\x^{(i)}_*$ refers to $\x_*$ is inferred with the $i$th submodel. 
Finally, denoting the means and variances of the local predictive distributions as $\hat{\mu}_i$ and $\hat{\sigma}^2_i$, respectively, we compute the aggregated statistics of the corrected PoE predictions as:
%\begin{align*}
%\mu_*^{\text{cpoe}} 
%&= (\sigma_*^{\text{cpoe}})^2 
%\bigg[ \sigma_{1}^{-2}(\x^{(1)}_*)\mu_{1}(\x^{(1)}_*)\\ 
%& %\phantom{(\sigma_*^{\text{cpoe}})^2 \bigg[ \frac{1}{I-1}\;}
%+ \sum_{i=2}^I\left( \sigma_i^{-2}(\x^{(i)}_*)\mu_i(\x^{(i)}_*) - \sigma_{1}^{-2}(\x^{(1)}_*)\mu_{1}(\x^{(1)}_*) \right)\bigg],\\
%(\sigma_*^{\text{cpoe}})^{-2} &= \sigma_{1}^{-2}(\x^{(1)}_*) +  \sum_{i=2}^I \left[\sigma_i^{-2}(\x^{(i)}_*) - \sigma_{1}^{-2}(\x^{(1)}_*) \right].
%\end{align*}
%
\begin{align*}
\mu_*^{\text{cpoe}} 
&= (\sigma_*^{\text{cpoe}})^2 
\bigg[ \hat{\sigma}_{1}^{-2}\hat{\mu}_{1}
+ \sum_{i=2}^I\left( \hat{\sigma}_i^{-2}\hat{\mu}_i - \hat{\sigma}_{1}^{-2}\hat{\mu}_{1} \right)\bigg],\\
(\sigma_*^{\text{cpoe}})^{-2} &= \hat{\sigma}_{1}^{-2} +  \sum_{i=2}^I \left[\hat{\sigma}_i^{-2} - \hat{\sigma}_{1}^{-2} \right].
\end{align*}

In their distributed GP framework, Deisenroth and Ng \shortcite{Deisenroth+Ng:2015} used a re-weighted variant of PoE,  which they coined the robust Bayesian committee machine (rBCM). Although rBCM has been shown to outperform the basic PoE for GP-regression, in our current setup, we have not observed any advantage of it over PoE. We have therefore formulated our framework using standard PoE but note that the extension to rBCM is straightforward.

\subsection{Improved Solution with Intermediate Aggregation}\label{sec:intermediate_aggregation}

%Aggregate submodels into a smaller number of submodels, on which we then do PoE. Since the aggregation already takes care of the over-counting of the first subset, we can use standard PoE for predictions. The aggregation of submodels into medium-sized models improves predictive performance. The aggregated models are small enough to enjoy the advantages of using PoE but too big to be learned as such.

% from the rebuttal:
%PoE effectively approximates the full-data covariance matrix with a block-diagonal covariance matrix, with block sizes corresponding to subset sizes. With larger blocks, PoE provides a closer approximation to the full covariance matrix, which can be expected to result in better accuracy. 
PoE aggregates predictions from local submodels learned on data subsets, effectively using a block-diagonal approximation of the full-data covariance matrix. 
With larger submodels, PoE provides a closer approximation to the full covariance matrix, which can be expected to result in better predictive accuracy. 
Here we introduce an \emph{intermediate aggregation} strategy, by which submodels are aggregated for improved predictive performance, while the initial training of submodels is still done on smaller subsets with lower computational cost. 
%The aggregation of submodels is facilitated by 
%Moreover, our 
While latent variable models are in general non-identifiable, making a direct aggregation of local models difficult to carry out in a meaningful way, the incremental learning introduced in Section~\ref{sec:propagation}, encourages identifiability among local models, alleviating the problem.
%, which facilitates the aggregation of multiple submodels into medium-sized models for improved predictive performance.

%the parallel submodels to target similar regions of the latent space, which facilitates the aggregation of multiple submodels into medium-sized models for improved predictive performance.
%
%In section~\ref{sec:propagation}, we employed incremental learning to encourage the parallel submodels to target the same regions of the latent space. The use of incremental learning would make the model parameters to be located in similar region of model space. This further motivates us to aggregate multiple submodels into medium-sized models for improved predictive performance. 
%

The aggregation of submodels involves (i) stacking together local variational distributions, which are assumed to be independent across subsets, (ii) concatenating the corresponding data subsets, and finally (iii) aggregating the hyperparameters of the models. The model parameters 
%$\boldsymbol{\theta}$ do not have prior distributions and 
can be approximated using suitable statistics (e.g. mode, median or mean) of the distributions. In our implementation, we use the mode to approximate the kernel and Gaussian noise variance parameters, and use averaging to estimate inducing variables.

Since the first subset $\Y^{(1)}$ is used multiple times through incremental learning, the corresponding variational distribution $q(\X^{(1)})$ is obtained through the following aggregation:
%by aggregation from all must be 
%a
%
%The variational distribution corresponding to the the first subset $\Y^{(1)}$
%The first step in aggregating submodels is to combine the variational distributions of the latent variables for the first subset, which has been used multiple times, and parameters $\boldsymbol{\theta}$ (including kernel parameters, inducing variables, and noise variance) 
%from different submodels to obtain an approximation of the model parameters. The aggregated variational approximation 
%$q(\X^{(1)})$ will obtained as follows:
\begin{align*}
	q\left(\X^{(1)}\right) &= \hat{q}_{1}\left(\X^{(1)}\right) \prod_{i=2}^I \left[ \hat{q}_i\left(\X^{(1)}\right) \hat{q}_1\left(\X^{(1)}\right)^{-1} \right], \\
	&= \prod_{n=1}^{N_{1}} \mathcal{N}\left(\x_{n,:}|\hat{\mu}^{*}_n,\hat{S}^{*}_n\right),
\end{align*}
where
\begin{align*}
	\left[\hat{S}^{*}_{n}\right]^{-1} & = \left[\hat{S}^{(1)}_{n}\right]^{-1} + \sum_{i=2}^I\left( \left[\hat{S}^{(i)}_{n}\right]^{-1} - \left[\hat{S}^{(1)}_{n}\right]^{-1}\right),\nonumber\\ 
	\hat{\mu}^{*}_n &=  \hat{S}^{*}_n
 \biggl[[\hat{S}^{(1)}_{n}]^{-1} \hat{\mu}^{(1)}_{n} + \notag \\
    &\mathrel{\phantom{=}} \sum_{i=2}^I \left( \left[\hat{S}^{(i)}_{n}\right]^{-1} \hat{\mu}^{(i)}_{n} - \left[\hat{S}^{(1)}_{n}\right]^{-1} \hat{\mu}^{(1)}_{n}\right)\biggr].\nonumber
\end{align*}
Above, each of the variational distributions $\hat{q}_i(\X^{(1)})$ is Gaussian, of the form given by Equation~(\ref{eq:variational_post}). 
%
%The model parameters $\boldsymbol{\theta}$ do not have prior distributions and can be approximated using some suitable statistics (e.g. mode, median or mean) of the distributions. In our implementation, we use the mode to approximate the kernel and Gaussian noise variance parameters, and use averaging to estimate inducing variables.

%After obtaining an approximation for $q(\X^{(1)})$ and the  model parameters $\boldsymbol{\theta}$, we can aggregate submodels to obtain medium-sized models by concatenating multiple subsets together. This involves using the approximated model parameters $\boldsymbol{\theta}$ for all aggregated models, and concatenating the variational distribution $q(\X^{(i)})$ of the latent variables of the corresponding subsets to get the variational distribution for the latent variables. 

Note that after intermediate aggregation, each training subset is used only once to make predictions, and we may use the ordinary PoE formulation in Equation (\ref{eq:poe_gp_regression}) for prediction.

\subsection{Computational Cost}\label{sec:scalability}

Our method aims to leverage the scaling properties of sparse GP for training and those of PoE for prediction. Thus, for data partitioned into subsets of size  $N_i$, $i=1,\ldots,I$,
and assuming that a sufficient number of parallel workers is available, 
the time complexity for training is $\mathcal{O}\left(\max_i(N_{i} M^2)\cdot D\right)$, where $M<N_i$ is the number of inducing points and $D$ reflects the fact that variational updates have to be performed separately for each dimension of sparsely observed data. For prediction, the cost is  $\mathcal{O}\left(\max_i(N_i^2)\right)$. For incremental learning and intermediate aggregation, $N_i$ refers to the size of the concatenation of multiple subsets. By intermediate aggregation of submodels, we are able to trade off prediction cost against accuracy.

\section{Experiments}\label{sec:experiments}

In this section, we evaluate the predictive performance of the proposed method for out-of-matrix prediction problems on simulated and real-world chemogenomic data, and compare it with two alternative approaches: (i) the embarrassingly parallel or subset of data (SoD) approach, which has been widely studied to scale up Gaussian Process regression models, and (ii) Macau, Bayesian multi-relational factorization with side information \cite{Simm+others:2015}, supporting out-of-matrix prediction. Macau is implemented with highly optimized C libraries, and is available for experiments on large-scale data. The comparison with the SoD approach shows the advantage of our method in sharing information among submodels, while the comparison with Macau shows the benefit of using Bayesian non-linear matrix factorization. We emphasize, however, that the choice and effectiveness of a model always depends on the problem at hand.

\paragraph{Simulated data.}
We generated synthetic data using non-linear signals corrupted with Gaussian noise, using matern data generator available in the  GPy\footnote{\label{GPy}\url{https://sheffieldml.github.io/GPy/}} software. The data has three views $\Y=\{\Y^{v}: v=1,2,3\}$, the dimension of the views is as follows: N = 25,000, $D^{1}$ = 150, $D^{2}$ = 100, $D^{3}$ = 150. As the task is to perform out-of-matrix prediction, we randomly selected 20\% of the rows as a test set, using the remaining rows as the training set. In addition, 80\% of the data in the first view were masked as missing, to simulate the sparsely observed target data in a real-world application. The other two views were regarded as side information and were fully observed. Unlike Bayesian GP-LVM, Macau cannot handle missing values in the side data.%\cite{gpy2014}

\paragraph{Real-world data.}
We performed the experiments on ExCAPE-DB data \cite{Sun+others:2017}, which is an aggregation of public compound-target bioactivity data and describes interactions between drugs and targets using the pIC50\footnote{IC50 (units in $\mu\text{M}$) is the concentration of drug at which 50\% of the target is inhibited. The lower the IC50 of the drug, the less likely the drug will be to have some off-target effect (e.g. potential toxicity) that is not desired. $\text{pIC50} = -\text{log}_{10}(\text{IC50})$.} measure. It has 969,725 compounds and 806 targets with 76,384,863 observed bioactivities. The dataset has 469 chem2vec features as side information which are generated from ECFP fingerprint features for the compounds using word2vec software. We used 3-fold cross validation to split the training and test set, where about 30\% of the rows or compounds were chosen as test set in each fold.

\subsection{Experimental Setup} 

%\paragraph{Experimental setting.} 
The experimental setting for MRD models is: number of inducing points 100, optimization through scaled conjugate gradients (SCG) with 500 iterations. For the SoD approach, the latent variables were initialized with PPCA method. We ran Macau with Gibbs sampling for 1200 iterations, discarded the first 800 samples as burn-in and saved every second of the remaining samples yielding in total 200 posterior samples. We set the dimension of latent variables K=10 for ExCAPE-DB data, K=5 for simulated data for all methods.

For the proposed and SoD methods, we partitioned the simulated data into 10x1 subsets and ExCAPE-DB data into 400x1 subsets. Other partitions are also possible; we have chosen the size of subsets such that Bayesian inference could be performed for the subsets in reasonable time on a single CPU. Notice that the views with missing values are generally sparsely observed in many real-world applications, which makes it challenging to learn informative models for such data. Following Qin et al. \shortcite{Qin+others:2019}, we reordered the rows and columns of training data in descending order according to the proportion of observations in them. This makes the first subset the most densely observed block, thus making the resulting submodel informative and facilitating the parallel inferences in the following stages.

%\subsection{Simulated data}
%\iffalse
\begin{table*}[!t]%[!htb]
  \centering
  \renewcommand{\arraystretch}{1.05}
{\fontsize{10pt}{10pt}\selectfont
 \begin{tabular}{l|c|c|c|c|c}
%\hline
\toprule
\multirow{2}{*}{Kernel}   &  \multirow{2}{*}{Macau} &  \multirow{2}{*}{SoD} &  \multicolumn{2}{c|}{Proposed methods} &   \multirow{2}{*}{Full posterior} \\  \cline{4-5}
  	&    &  &   PoE &  Intermediate aggregation   &   \\ %\hline 
\midrule
\multicolumn{6}{l}{RMSE: the smaller, the better.}  \\  \hline

Linear   &  0.8927 $\pm$ 0.010 & 0.747 $\pm$ 0.034& 0.685 $\pm$ 0.041& \textbf{0.656 $\pm$ 0.038} & \textbf{0.656 $\pm$ 0.039} \\  \hline  
RBF      &  -  &  0.825 $\pm$ 0.034& 0.791 $\pm$ 0.048& 0.683 $\pm$ 0.038 & \textbf{0.658 $\pm$ 0.039} \\  \hline  
Matern32 &  - &  0.824 $\pm$ 0.032& 0.772 $\pm$ 0.048& 0.687 $\pm$ 0.039 & \textbf{0.656 $\pm$ 0.038} \\  %\hline  

\midrule
\multicolumn{6}{l}{Spearman correlation score: the larger, the better.}  \\  \hline  

Linear   &  0.6971 $\pm$ 0.044 &  0.713 $\pm$ 0.056& 0.726 $\pm$ 0.048& \textbf{0.744 $\pm$ 0.044} & \textbf{0.744 $\pm$ 0.045} \\  \hline  
RBF      &  -  &  0.689 $\pm$ 0.060& 0.651 $\pm$ 0.080& 0.721 $\pm$ 0.048 & \textbf{0.742 $\pm$ 0.045} \\  \hline  
Matern32 &  - &  0.684 $\pm$ 0.064& 0.672 $\pm$ 0.083& 0.718 $\pm$ 0.047& \textbf{0.744 $\pm$ 0.044} \\  %\hline  

\midrule
\multicolumn{6}{l}{Wall-clock time (secs.)}  \\  \hline
%Train time(secs) per method on simulated data
%Linear   &  171.44   &  26182.63 & 52929.52 & 52929.52 & 245606.54  \\  \hline  
%RBF      &  -  & 28550.93 & 49573.35 & 49573.35 & 165597.31   \\  \hline  
%Matern32 &  - &  34322.34 & 64057.95 & 64057.95 & 97705.28  \\  %\hline  
%
%%Prediction time (secs) per batch with 2500 data points on simulated data
%Linear	& -  & 1548.118	& 2261.776	& 4125.507	& 4176.427  \\  \hline  
%Rbf	&  - & 2820.690	& 3638.255	& 7733.099	& 10478.152  \\  \hline  
%Matern32  & - &  2434.954	& 3139.188	& 7887.872	& 8597.736  \\  \hline  

%Total wall-clock time (train time + predict time per test batch)
Linear    &  171.44   & 27730.748 & 55191.296  &  57055.027 & 249782.967  \\  \hline  
RBF	  & -  &  31371.620 & 53211.605  & 57306.449 & 176075.462  \\  \hline  
Matern32  & -  &  36757.294 & 67197.138  & 71945.822 & 106303.016  \\  %\hline  
\bottomrule
\end{tabular}
}
 \caption{Comparison of performance metrics for different methods on simulated data. The results are averaged over 5 folds.}% Size of data: N=25000, D1=150, D2=100, D3=150. % RMSE, spearman correlation and wall-clock time
\label{tab:simulated_data}
\end{table*}

%\fi

%%%%%%%% commented block: %%%%%%%%%%%%%%%%%%%%%%%%%%%%
%%%Note: results for performance metrics on ExCAPE-DB, averaged over only hit target queries for each fold%%%%%%%%%%
%\iffalse

\begin{table*}[!t]%[!htb]
\renewcommand{\arraystretch}{1.05}
\centering
{\fontsize{9pt}{9pt}\selectfont
\begin{tabular}{clccccc}
\toprule
Affinity   &   \multirow{2}{*}{Model}  &  \multirow{2}{*}{RMSE}    &   \multirow{2}{*}{F1-score}   &  AUC-ROC   &   Ratio of successful    &   Wall-clock \\ 
level   &      &     &      &  score   &   queries   &  time (secs.) \\ 

\midrule

5   &   Macau   &   1.108 $\pm$ 0.069   &   0.886 $\pm$ 0.003   &   0.805 $\pm$ 0.009   &   0.319 $\pm$ 0.024   &   37041.8 \\ 
5   &   SoD   &   0.914 $\pm$ 0.023   &   0.890 $\pm$ 0.011   &   0.791 $\pm$ 0.003   &   0.363 $\pm$ 0.006   &   63110.16 \\ 

\multicolumn{7}{l}{Proposed methods:} \\

5   &   PoE   &   0.831 $\pm$ 0.021   &   0.900 $\pm$ 0.001   &   0.805 $\pm$ 0.002   &   0.309 $\pm$ 0.008   &   92419.06 \\ 
5   &   Intermediate aggregation (nAggs=10)   &   0.743 $\pm$ 0.009   &   \textbf{0.919 $\pm$ 0.005}   &   0.811 $\pm$ 0.006   &   0.405 $\pm$ 0.018   &   93331.23 \\ 
5   &   Intermediate aggregation (nAggs=20)   &   \textbf{0.736 $\pm$ 0.004}   &   0.914 $\pm$ 0.003   &   \textbf{0.813 $\pm$ 0.004}   &   \textbf{0.455 $\pm$ 0.008}   &   100492.9 \\ 
\hline

6   &   Macau   &   1.123 $\pm$ 0.065   &   0.783 $\pm$ 0.013   &   0.799 $\pm$ 0.006   &   0.318 $\pm$ 0.003   &   37041.8\\ 
6   &   SoD   &   0.930 $\pm$ 0.021   &   0.787 $\pm$ 0.011   &   0.791 $\pm$ 0.005   &   0.381 $\pm$ 0.019   &   63110.16\\ 

\multicolumn{7}{l}{Proposed methods:} \\

6   &   PoE   &   0.837 $\pm$ 0.022   &   0.846 $\pm$ 0.011   &   0.811 $\pm$ 0.003   &   0.285 $\pm$ 0.004   &   92419.06\\ 
6   &   Intermediate aggregation (nAggs=10)   &   \textbf{0.775 $\pm$ 0.028}   &   \textbf{0.851 $\pm$ 0.015}   &   \textbf{0.817 $\pm$ 0.003}   &   0.376 $\pm$ 0.025   &   93331.23\\ 
6   &  Intermediate aggregation (nAggs=20)   &   0.789 $\pm$ 0.006   &   0.838 $\pm$ 0.005   &   0.816 $\pm$ 0.004   &   \textbf{0.434 $\pm$ 0.016}   &   100492.9\\ 

\bottomrule
\end{tabular}
}
\caption{Comparison of RMSE, F1-score, AUC-ROC  score and the ratio of successful queries (i.e. queries with AUC-ROC score larger than 0.7 for the targets) for out-of-matrix prediction on ExCAPE-DB by different methods. The first three metrics are calculated for only the successful queries, the ratio is defined as \#successfulQueries / \#validQueries, where a query target is considered to be valid if it has at least 100 observed bioactivity, at least 25 active and 25 inactive compounds. The results are averaged over 3 runs.}\label{tab:excape_db_data}
\end{table*}

% total wall-clock time = train time + Prediction time (secs) per batch with 2500 data points on ExCAPE-DB dataset
%Method & Train time & Prediction time & Total wall-clock time
%LIS	& 63010.16 & 100 & 63110.16
%PoE	& 91919.06 & 500 &  92419.06
%Intermediate aggregation (nAggs=10)	& 91919.06 & 1412.170 & 93331.23
%Intermediate aggregation (nAggs=20)	& 91919.06 & 8573.850 & 100492.9

%\fi
%%%%%%%%%%%%%%%%%%%%%%%%%%%%%%%%%%%%%%%%%%%%%%%%%%%%%%

%\paragraph{Performance metrics.} 
We evaluated performance by predictive accuracy and the quality of prediction for downstream ranking tasks. Root mean squared error (RMSE) is a common performance measure for matrix completion. In real-world applications, such as item recommendation or drug discovery, we are more interested in the performance of the ranking task, for instance how many of the recommended items the user actually clicks or buys, how many drugs recommended by models actually have the desired effect for the disease. For this purpose, we regard matrix completion as a classification task (of whether a prediction is relevant or not at a given threshold), use F1- and AUC-ROC score as performance metrics for ExCAPE-DB. Furthermore, following Qin et al. \shortcite{Qin+others:2019}, we use the wall-clock time\footnote{Wall-clock time measures the real time between the start and the end of a program. For parallel processes, we use the wall-clock time of the slowest process.} to measure the speed-up achieved by parallelization. For our method, the reported wall-clock time is calculated by summing the maximum wall-clock times of submodels for each inference stage plus the wall-clock time of making prediction. %plus the wall-clock time of the intermediate aggregation step.

For compound activity prediction tasks, we use a pIC50 cutoff (a.k.a. affinity level) at 5 and 6, corresponding to concentrations of 10$\mu\text{M}$ and 1$\mu\text{M}$, respectively. The test set was further filtered by only keeping targets having at least 100 compounds, at least 25 active compounds, and 25 inactive compounds, to ensure a minimum number of positive and negative data points.

Macau\footnote{We ran the Macau version available in SMURFF software: \url{https://github.com/ExaScience/smurff}.} was run on compute nodes with 20 CPUs; all the other methods were run on a single CPU. Our implementation is based on the GPy$^{\ref{GPy}}$ package.

\subsection{Results}

The results for simulated and ExCAPE-DB data are given in Table~\ref{tab:simulated_data} and \ref{tab:excape_db_data}, respectively. In Table~\ref{tab:simulated_data}, column `Full posterior' refers to the performance of MRD learned from the full data; column `Intermediate aggregation' refers to our method which works by first aggregating multiple submodels into a model with reasonable size (as long as the compute node can still accommodate the model to make predictions) and then perform predictions by aggregating predictions from multiple experts with PoE.

It is clear from Table~\ref{tab:simulated_data} that the model with full posterior performs better than other methods in terms of predictive performance; our intermediate aggregation method achieves competitive results while being much faster than the full posterior. The intermediate aggregation method also performs better than the SoD approach and the variant of our method without the intermediate aggregation step. With a linear kernel, all MRD models perform better than Macau.

For ExCAPE-DB data, our intermediate aggregation method (by aggregating 10 or 20 submodels to obtain larger models for prediction) performs much better than all the other methods in all performance metrics for different affinity levels.  At both affinity levels, all versions of our proposed method perform better than the SoD method in terms of RMSE, F1-score and AUC-ROC score. Again, we observed that all MRD methods perform better than Macau in all performance metrics. In both tables, the wall-clock times of our methods are larger than that of the SoD approach. This is due to the two-stage parallel inferences of the proposed scheme.

To summarise, the proposed method with an intermediate aggregation step achieves a better trade-off between predictive accuracy and computation time. The proposed method performs better than the embarrassingly parallel approaches for scalable Gaussian process models and a state-of-the-art highly optimized implementation of linear Bayesian matrix factorization with side information.

\section{Conclusion}\label{sec:conclusion}

%Merits and implications of our method. 
In this paper, we have introduced a scalable approach for Bayesian non-linear matrix completion. 
%To the best of our knowledge this is the first fully Bayesian approach for non-linear matrix completion. 
We have argued that a key factor in constructing distributed solutions for massive-scale data is to limit the communication required between computational units. 
To this end, we have introduced a computational scheme which 
leverages embarrassingly parallel techniques developed for Gaussian process regression 
%but avoids problems associated with combining non-identifiable submodels. 
by suitably adapting them for Bayesian Gaussian process latent variable models. 
The resulting framework is almost as communication-efficient as embarrassingly parallel computation, adding only one additional stage of communication, while achieving accuracy close to the non-distributed full data solution.  

\section*{Acknowledgements}
The authors gratefully acknowledge the computational resources provided by the Aalto Science-IT project and support by the Academy of Finland (Finnish Center for Artificial Intelligence, FCAI, and projects 319264, 292334).

\newpage

\bibliographystyle{named} % natbib
\bibliography{distributedMRD}

\end{document}